\documentclass[a4paper,twoside]{article}

\usepackage[utf8]{inputenc}
\usepackage[english]{babel}

\usepackage{epsfig}
\usepackage{subfigure}
\usepackage{calc}
\usepackage{amssymb}
\usepackage{amstext}
\usepackage{amsmath}
\usepackage{amsthm}
\usepackage{multicol,multirow}
\usepackage[autolanguage]{numprint}
\usepackage{pslatex}
\usepackage{apalike}
\usepackage{SciTePress}
\usepackage[small]{caption}

\usepackage{rotating}
\usepackage{xfrac}
\usepackage{tikz}
\tikzstyle{every node}=[font=\small]
\tikzset{hiddenstate/.style={circle,text width=12pt,draw=black,align=center}}
\tikzset{obsstate/.style={circle,text width=12pt,draw=black,fill=black!25,align=center}}

\usepackage{todonotes}
\usepackage{import}
\usepackage{acronym}

\def\latinstyle{\textit}
\newcommand{\latin}[1]{\latinstyle{#1}}
\newcommand{\ie}{\latin{i.e.}}
\newcommand{\eg}{\latin{e.g.}}

\acrodef{ACP}{Analyse en Composantes Principales}
\acrodef{PCA}{Principal Component Analysis} 
\acrodef{ACI}{Analyse en Composantes Indépendantes}
\acrodef{ICA}{Independant Component Analysis} 
\acrodef{ACC}{Analyse en Composantes Curvilinéaire}
\acrodef{CCA}{Curvilinear Component Analysis}

\acrodef{DoG}{Difference of Gaussians}

\acrodef{CBIR}{Content Based Image Retrieval}

\acrodef{SIFT}{Scale-Invariant Feature Transform}
\acrodef{SURF}{Speeded-Up Robust Features}
\acrodef{HOG}{Histogram of Oriented Gradients}
\acrodef{RHOG}[R-HOG]{Rectangular Histogram of Oriented Gradients}
\acrodef{CHOG}[C-HOG]{Circular Histogram of Oriented Gradients}
\acrodef{CENTRIST}{CENsus TRansform hISTogram}
\acrodef{LBP}{Local Binary Pattern}
\acrodef{CT}{Census Transform}
\acrodef{MSER}{Maximally Stable Extremal Region}
\acrodef{RBC}{Recognition by components}
\acrodef{GSD}{Geon Structural Description}
\acrodef{DST}{Discriminant Spectral Template}
\acrodef{SPM}{Spatial Pyramid Matching}

\acrodef{ERM}{Empirical Risk Minimization}
\acrodef{SRM}{Structural Risk Minimization}
\acrodef{WCA}{Worst Case Analysis}
\acrodef{TSUC}{Two-Sided Uniform Convergence}
\acrodef{OSUC}{One-Sided Uniform Convergence}

\acrodef{BoW}{Bag-of-Words}
\acrodef{tBoW}{temporal Bag-of-Words}
\acrodef{BFMC}{Bayesian Filtering with Markov Chains}
\acrodef{SOM}{Self-Organizing Map}
\acrodef{BMU}{Best-Matching Unit}
\acrodef{CBN}{Classificateur Bay\'{e}sien Na\"{i}f}
\acrodef{NBC}{Naive Bayes Classifier}
\acrodef{ADL}{Analyse Discriminante Linéaire}
\acrodef{FDA}{Fisher Discriminant Analysis}
\acrodef{WTA}{Winner-take-all}
\acrodef{MMO}{Mod\`{e}le de Markov Observable}
\acrodef{OMM}{Observable Markov Model}
\acrodef{MMC}{Modèle de Markov Caché}
\acrodef{HMM}{Hidden Markov Model}
\acrodef{HOHMM}{High-Order Hidden Markov Model}
\acrodef{SVM}{Support Vector Machines}
\acrodef{LVQ}{Learned Vector Quantization}
\acrodef{TFIDF}[tf-idf]{Term Frequency - Inverse Document Frequency}
\acrodef{kPPV}[$k$-PPV]{$k$ plus proches voisins}
\acrodef{kNN}[$k$-NN]{$k$ nearest neighbors}
\acrodef{pLSA}{Probabilistic Latent Semantic Analysis}
\acrodef{LDA}{Latent Dirichlet Allocation}

\acrodef{COLD}{COsy Localization Database}
\acrodef{VPC}{Visual Place Categorization}

\acrodef{GMM}{Gaussian Mixture Model}
\acrodef{EM}{Expectation-Maximization}

\acrodef{CFR}{Caractéristique de Fonctionnement du Récepteur}
\acrodef{TVP}{Taux de Vrais Positifs} 
\acrodef{TVN}{Taux de Vrais Négatifs} 
\acrodef{TFP}{Taux de Faux Positifs} 
\acrodef{ROC}{Receiver Operating Characteristic}
\acrodef{TPR}{True Positive Rate}  
\acrodef{TNR}{True Negative Rate}  
\acrodef{FPR}{False Positive Rate} 

\acrodef{SLAM}{Simultaneous Location And Mapping}
\acrodef{CLM}{Concurrent Location and Mapping}

\usepackage{amsmath, amssymb, amsthm, textcomp, stmaryrd}
\usepackage{textcase}

\def\setextleftdelim{\lbrace}
\def\setextrightdelim{\rbrace}
\newcommand{\setext}[1]{\ensuremath{\left \setextleftdelim {#1} \right \setextrightdelim}} 

\def\setstyle{\mathcal}
\newcommand{\set}[1]{\ensuremath{\setstyle{#1}}}

\def\functionstyle{\mathrm}
\newcommand{\func}[1]{\ensuremath{\functionstyle{#1}}}
\newcommand{\apply}[2]{\ensuremath{#1 \left ( {#2} \right )}}

\newcommand{\vectorstyle}[1]{\boldsymbol{#1}}
\renewcommand{\vector}[1]{\ensuremath{\vectorstyle{#1}}}

\newcommand{\RV}[1]{\ensuremath{\MakeTextUppercase{#1}}}
\newcommand{\cond}[2]{{#1} | {#2}}

\def\probSymbol{P}
\newcommand{\prob}[1]{\ensuremath{\apply{\func{\probSymbol}}{#1}}}
\newcommand{\condprob}[2]{\ensuremath{\prob{\cond{#1}{#2}}}}

\def\pdfSymbol{p}
\newcommand{\pdf}[1]{\ensuremath{\apply{\func{\pdfSymbol}}{#1}}}
\newcommand{\condpdf}[2]{\ensuremath{\pdf{\cond{#1}{#2}}}}

\newcommand{\stateSymbol}{x}
\newcommand{\state}{\vector{\stateSymbol}}
\newcommand{\obsSymbol}{z}
\newcommand{\obs}{\vector{\obsSymbol}}
\newcommand{\controlSymbol}{u}
\newcommand{\control}{\vector{\controlSymbol}}

\newcommand{\stateRVSymbol}{\RV{\stateSymbol}}
\newcommand{\stateRV}{\vector{\stateRVSymbol}}
\newcommand{\obsRVSymbol}{\RV{\obsSymbol}}
\newcommand{\obsRV}{\vector{\obsRVSymbol}}

\newcommand{\val}[2]{\ensuremath{{#1}_{#2}}}
\newcommand{\seq}[3]{\ensuremath{{#1}_{{#2}:{#3}}}}

\def\bibpath{.}

\acrodef{NLP}{Natural Language Processing}
\def\somSize{S}

\def\nWords{K}
\def\sampleRate{s}
\def\compressRate{m}

\begin{document}

\title{Using $n$-grams models for visual semantic place recognition}

\author{\authorname{%
Mathieu Dubois\sup{1,2}, Emmanuelle Frenoux\sup{1,2} and Philippe Tarroux\sup{2,3}}%
\affiliation{\sup{1}Univ Paris-Sud, Orsay, F-91405}%
\affiliation{\sup{2}LIMSI-CNRS, B.P. 133, Orsay, F-91403}%
\affiliation{\sup{3}\'{E}cole Normale Supérieure, 45 rue d'Ulm, Paris, F-75230}%
\email{\{mathieu.dubois, emmanuelle.frenoux, philippe.tarroux\}@limsi.fr}%
}


\abstract{
 The aim of this paper is to present a new method for visual place recognition.
 Our system combines global image characterization and visual words, which allows to use
efficient Bayesian filtering methods to integrate several images.
 More precisely, we extend the classical \acs{HMM} model with techniques inspired by the field of \acl{NLP}.
 This paper presents our system and the Bayesian filtering algorithm.
 The performance of our system and the influence of the main parameters are evaluated on a standard database.
 The discussion highlights the interest of using such models and proposes improvements.
}

\onecolumn \maketitle \normalsize \vfill

\section[Introduction]{\uppercase{Introduction}}
\noindent Semantic mapping (see~\cite{Nuchter2008}) is a relatively new field in robotics which aims to give the robot a high-level, human-compatible, understanding of its environment
in order to ease the integration of robots in daily environments, notably homes or workplaces.
Such environments are usually composed of discrete places which correspond to different functions.
For instance a house is usually made of different rooms and corridors used to move between them.
Such places are called \emph{semantic places} because they are defined in high-level human concepts as opposed to traditional low-level landmarks used in robot mapping.

In this context, it's important for the robot to be able to recognize in which place or category of places it lies.
Those tasks are called respectively instance recognition and categorization.
Semantic place recognition is then an important component of semantic mapping.
Moreover the semantic category of a place can be used to foster object detection and recognition
(giving priors on objects identity, location and scale) and to provide qualitative localization.

Different types of sensors have been employed for semantic place recognition.
The first works in this domain used range sensors to discriminate places based on geometrical information.
However the spatial configuration of two places of the same category (\eg{} two kitchens) can be very different.
Therefore geometrical information may not be useful for categorization.
Vision is the modality of choice for semantic place recognition because
it gives access to rich, allothetic information.
Although there are multimodal approaches, our work focuses on \emph{visual} place recognition.

In this article we will further develop an analogy between semantic place recognition and language modelling.
This analogy allows to design efficient temporal integration methods \ie{} to take several images into account in order to reduce ambiguity.
More precisely, we will extend the \acf{HMM} formalism with $n$-grams models.
Those models have been extensively used in \ac{NLP} 
and efficient estimation techniques have been proposed.
This paper aims to assess the use of such models in semantic place recognition.
The goal is to compare this temporal integration method to previously proposed models.
In particular we will study the influence of the length of the $n$-gram model and estimation procedure on performance.

The article is structured as follows.
Section~\ref{sec:related_works} presents related work.
Our model and its links with language modelling are described in section~\ref{sec:model_presentation}.
Section~\ref{sec:experimental_results} presents our experiments and the results.
Finally we conclude in section~\ref{sec:conclusion}.

%
%
%

\section[Related work]{\uppercase{Related work}} \label{sec:related_works}

\noindent Some authors (see~\cite{Vasudevan2007c}) use an object-based approach.
In this case they employ a standard algorithm for object localization and recognition.
Places are described by the frequency of objects found in them combined with constraints on their position.
However, object categorization is still a difficult task and the position of objects can greatly vary from one environment to another.
Therefore those approaches have not been used on large databases.

The vast majority of research on place recognition use techniques developed for visual scene classification.
We can distinguish methods using global features (see~\cite{h_Torralba2003}) and methods using descriptors computed around interest points (see~\cite{Ullah2008}).
\cite{Filliat2008} uses the \ac{BoW} model: local features are first clustered into a so-called dictionary of visual words learned by mean of a vector quantization algorithm.
An image is represented by the distribution of visual words found in it.
The major advantage is that the learning space is discretized but all geometrical information is lost.


Generally speaking using a single image or a single type of information is not enough for place recognition tasks.
Therefore a lot of research has been conducted to disambiguate perception.
\cite{Pronobis2007} use a confidence criterion to iteratively compute several cues from the same image until confidence in the classification is sufficiently high (or no more cues are available).

Another method to reduce ambiguity is to use several images to mutually disambiguate perception.
In \cite{Pronobis2010}, the authors use a simple spatio-temporal
accumulation process to filter the decision of a discriminative confidence-based place recognition system (which uses only one image to recognize the place).
One problem with this method is that the system needs to wait some time before giving a response.
Also, special care must be taken to detect places boundaries and to adjust the size of the bins.
\cite{h_Torralba2003} use a \ac{HMM} where each place is a hidden state and the feature vector stands for the observation.
The drawback is that the input space is continuous and high-dimensional.
The learning procedure is then computationally expensive.
\cite{Ranganathan2010} uses a technique called \emph{Bayesian online change-point detection}.
The main idea is to detect abrupt changes in the parameters of the input's statistics caused by moving from one place to another.
The main advantage is that the robot is able to learn in an unsupervised way but relies on the hypothesis that the shape of the distribution is the same for every place.

Several works (see~\cite{Wu2009,Guillaume2011,Dubois2011}) have combined global image description and vector quantization.
In this case, each image is described by a single visual word.
The sequence of images is then translated into a sequence of words.
Such techniques allow to draw a parallel between place recognition and language modelling.
\cite{Wu2009} propose to use a \ac{HMM} with discretized signatures.
Temporal integration is performed with Bayesian filtering (see section~\ref{sec:model_presentation}).
\cite{Dubois2011} propose to use an extended model called auto-regressive~\ac{HMM} to take into account the dependence between images.

In this paper we push this idea a step further.
The next section presents our models and its relations to the standard~\ac{HMM} model.

%
%

\section{PLACE RECOGNITION WITH $n$-GRAMS} \label{sec:model_presentation}

\noindent Our model is similar to the one described in~\cite{Guillaume2011,Dubois2011}.
Each image is described by a unique feature vector which is mapped to a given visual word thanks to a vector quantization algorithm (see section~\ref{subsec:image_lvq}).
The main novelty lies in the use of High-Order Hidden Markov Model (see section~\ref{subsec:hohmm})
and techniques for visual word selection (see~\ref{subsec:word_selection}).

\subsection{High-Order Hidden Markov model} \label{subsec:hohmm}

\begin{figure*}[htpb]
 \centering
 \subfigure[]{\begin{tikzpicture}[shorten >=1pt,->]
      \foreach \t/\x in {{t-1}/0, {t}/2}
	\node[hiddenstate] (\stateSymbol-\t) at (\x,1.5) {$\val{\stateRV}{\t}$};
      \foreach \from/\to in {{t-1}/{t}}
	\draw (\stateSymbol-\from) -- (\stateSymbol-\to);
	
	
	
      \foreach \t/\x in {{t-1}/0, {t}/2}
	\node[obsstate] (\obsSymbol-\t) at (\x,0) {$\val{\obsRV}{\t}$};

      \foreach \t in {{t-1}, {t}}
	\draw (\stateSymbol-\t) -- (\obsSymbol-\t);
\end{tikzpicture}\label{fig:dbn_probabilistic_rob}} \quad
 \subfigure[]{\begin{tikzpicture}[shorten >=1pt,->]
      \foreach \t/\text/\x in {{t-l}/{t-\ell}/-2, {t-1}/{t-1}/0, {t}/{t}/2}
	\node[hiddenstate] (\stateSymbol-\t) at (\x,1.5) {$\val{\stateRV}{\text}$};
      \node at (-1,1.5) {$\ldots$};
      \foreach \from/\to/\control in {{t-l}/t/{.. controls +(+30:2cm) and +(+150:1cm) ..}, {t-1}/t/{--}}
        \draw (\stateSymbol-\from) \control (\stateSymbol-\to);
	
%
%
      \foreach \t/\text/\x in {{t-n}/{t-n}/-4, {t-2}/{t-2}/-2, {t-1}/{t-1}/0, {t}/{t}/2}
	\node[obsstate] (\obsSymbol-\t) at (\x,0) {$\val{\obsRV}{\t}$};
      \node at (-3,0) {$\ldots$};

      \draw (\stateSymbol-t) -- (\obsSymbol-t);

      \foreach \from/\to/\control in {{t-n}/t/{.. controls +(+30:2cm) and +(+150:1cm) ..}, {t-2}/t/{.. controls +(+30:2cm) and +(+150:1cm) ..}, {t-1}/t/{--}}
        \draw (\obsSymbol-\from) \control (\obsSymbol-\to);
\end{tikzpicture}\label{fig:hohmm}}
 \caption{\subref{fig:dbn_probabilistic_rob} The classical \ac{HMM} model. \subref{fig:hohmm} The \ac{HOHMM} model (we only show nodes that have an influence on $\val{\state}{t}$ and $\val{\obs}{t}$).}
\end{figure*}
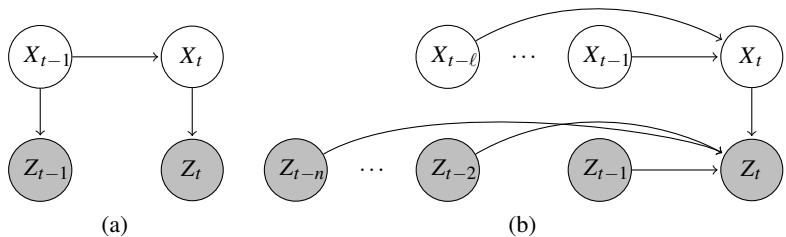

In~\acp{HMM} the relationship between
$\val{x}{t}$, the robot's knowledge of the world at time $t$, and $\val{\obs}{t}$, its perception is represented by figure~\ref{fig:dbn_probabilistic_rob}.
In the case of place recognition, the state is a discrete random variable which represents the place the robot is in at time $t$.
In this model, each place $c_i \in \set{C}$ is modelled by the continuous probability distribution $\condpdf{\val{\obs}{t}}{\val{\state}{t} = c_i}$.
This formalism allows to efficiently estimate the a posteriori probability $\operatorname{bel}(\val{\state}{t})=\condprob{\val{\state}{t}}{\seq{\obs}{1}{t}}$ by a recursive equation (see~\cite{Wu2009})
given the discrete place transition probability distribution $\condprob{\val{\state}{t}}{\val{\state}{t-1}}$ which encodes the topology of the environment.

It is assumed that the current observation depends only on the current hidden state \ie{} that the state is complete.
However, there is a huge semantic gap between the human notion of a place and what can be extracted from an image.
Several authors have proposed extensions of the classic~\ac{HMM} to take into account long-term dependencies between observations (see~\cite{Berchtold2002,Lee2006}).
In this paper we will call this model~\ac{HOHMM}.
In this case, the current knowledge $\val{\state}{t}$ depends on the last $\ell$ states $\seq{\state}{t-\ell}{t-1}$.
Similarly the current observation $\val{\obs}{t}$ depends on $\val{\state}{t}$ and the $n$ previous observations $\seq{\obs}{t-n}{t-1}$ (see figure~\ref{fig:hohmm}).
In this paper we restrict ourselves to the case $\ell=1$.
Therefore the state transition matrix is unchanged.

The a posteriori distribution $\operatorname{bel}\left(\val{\state}{t}\right)$ is given by:
\begin{equation}\label{eq:bfhohmm}
 \operatorname{bel}\left(\val{\state}{t}\right) = \condpdf{\val{\obs}{t}}{\seq{\obs}{t-n}{t-1},\val{\state}{t}} \sum_{c_{i} \in \set{C}} \condprob{\val{\state}{t}}{\val{\state}{t-1}}\operatorname{bel}\left(\val{\state}{t-1}\right)
\end{equation}
The place model is given by the distribution $\condpdf{\val{\obs}{t}}{\seq{\obs}{t-n}{t-1},\val{\state}{t} = c_i}$.
This probability distribution may be very difficult to learn because it is continuous.

\subsection[\ac{HOHMM} and Visual Words]{\ac{HOHMM} and Visual Words}


In order to simplify learning of the place model~\cite{Guillaume2011,Dubois2011} have proposed to use global image characterization in combination with vector quantization algorithms to discretize them.
In this case the variable $\val{\obs}{t}$ is reduced to a discrete random variable with a finite number of values $\setext{1,\dots, \nWords}$
where $\nWords$ is the number of words in the dictionary.

In this case, the model of place $c_i$ is given by the discrete probability distribution $\condprob{\val{\obs}{t}}{\seq{\obs}{t-n}{t-1},\val{\state}{t} = c_i}$.
In~\ac{NLP}, such a model is known as a $n+1$-gram model because it uses $n+1$ words.
\cite{h_Chen1996} have shown that the estimation of the model from empirical data is an important factor.
One problem is that even with a large training set, some sequences of words will not be observed in training data for a given class and therefore they will be assigned a null probability in this class' model.
If such a sequence is observed in the testing set then the a posteriori probability of this class will be clamped to $0$ due to equation~\ref{eq:bfhohmm}.
To avoid this problem, it is necessary to take some probability mass from the observed sequences and distribute it to unobserved sequences.
Those techniques are called smoothing or discounting.
We refer the reader to~\cite{Manning1999} for a unified presentation of smoothing techniques.
We use the SRILM toolkit to learn the $n$-grams models.


\subsection{Image Characterization and Vector Quantization} \label{subsec:image_lvq}

%

To characterize the images we use the GIST descriptors (see~\cite{h_Torralba2003}) which is an efficient global image characterization.
The image is divided into $4 \times 4$ subwindows (we use only the luminance channel) and filtered using a bank of Gabor filters (we use 4 scales and 6 orientations).
The energy of the filter is then averaged on each subwindow for each scale and orientation.
Finally the output is projected on the first $80$ principal components which explains more than 99\% of the variance.
Thus this descriptor captures the most significant spatial orientations at a given scale.

The vector quantization algorithm used in this paper is the~\ac{SOM} (see~\cite{h_Kohonen1990}).
In the current set-up the training of the \ac{SOM} is performed off-line on a set of randomly chosen images made of $\sfrac{1}{3}$ of the \acs{COLD} DB.
The number of neurons on the map sets the number of words in the visual dictionary which is an important parameter of the system.
We use square maps parametrized by their length $\somSize$ (therefore $\nWords = \somSize^2$).
In this paper we will use $\somSize = 10$ and $\somSize = 20$.
Those values were selected because it has been shown that small maps have a good performance on categorization tasks
while larger maps perform well for instance recognition (see~\cite{Guillaume2011}). 
Because the training algorithm is stochastic, the results vary from one~\ac{SOM} to another.
Therefore for each size $\somSize$, the results are averaged for $5$ \acp{SOM}.

\subsection{Visual Words Selection} \label{subsec:word_selection}

The sampling rate of most databases is several Hertz.
In this case, image at time $t+1$ is not very different from image at time $t$ and there is a high probability that they are described
by close vectors and therefore by the same visual word.
While this is a desirable feature of image description and vector quantization, this may be a problem for our method because the probability of seeing the same visual word than before will be very high. 
Therefore it might be interesting to use only a subset of the images (and then the words) for learning.

In order to evaluate this phenomenon we have computed the average number of consecutive time-steps which are characterized
by the same visual word for the training sequence used in section~\ref{sec:experimental_results}.
Results are given in table~\ref{tab:av_time_same_word}.

We will test three different strategies for selecting visual words.
The first one is simply to sub-sample the input image \ie{} to select 1 image out of $\sampleRate$ ($\sampleRate$ is the sub-sampling rate).
This strategy will be called ``subsample''.
The second strategy is to replace every sequence of $\compressRate$ identical prototypes by a unique instance of this word ($\compressRate$ is the compression rate).
We will call this strategy ``compress''.
The last strategy is to use the word at time $t$ only if it is different than the word at time $t-1$.
We will call this strategy ``unique''.
Those strategies are simple and can be implemented online on a real robot with limited computational power.


\begin{table}[htp]
 \centering
\begin{tabular}{|c|c|} \hline
 $\somSize$ & $\bar{t}$       \\ \hline
    10      & \numprint{3.15} \\ \hline
    20      & \numprint{2.67} \\ \hline
 \end{tabular}
 \caption{Average number of consecutive images represented by the same visual word for different \ac{SOM} size.}
 \label{tab:av_time_same_word}
\end{table}

In the next section we will present the experiment we carried out to study the use of this model for semantic place recognition.

%
%

\section{\uppercase{Experimental results}} \label{sec:experimental_results}

\subsection{Experimental Design}

\noindent We use the \acs{COLD} database (see~\cite{Ullah2008}) a standard database to evaluate vision-based place recognition systems.
It consists of sequences acquired by a human-driven robot in different laboratories across Europe under different illumination conditions (night, cloudy, sunny).
In each laboratory, two paths were explored (standard and extended).
Each path was followed at least 3 times under each illumination condition.
All the experiments were carried out with the perspective images.

Protocols proposed by~\cite{Ullah2008} uses only a few hundreds images per place which is not enough to robustly estimate the transition probabilities.
Therefore we designed a new experiment to evaluate the interest of our method.
We use only images acquired in Saarbruecken part B because other parts of the database are known to contain errors (\eg{} missing places or labellisation errors) or are not complete (\eg{} only one path was followed).
There are five classes (see table~\ref{tab:n_images}).
Training is performed with sequences number 1 and 2 from all the three illumination conditions.
Similarly, testing is performed with sequence 3 from all the illumination conditions.

\begin{table}[htbp]
\centering
\begin{scriptsize}
\begin{tabular}{c|c|c|c|c|c|} \cline{2-6}
                                & Office & Corridor & Printer area & Toilets & Kitchen \\\hline
 \multicolumn{1}{|c|}{Training} & \numprint{1375} & \numprint{4464} & \numprint{1190} & \numprint{3272} & \numprint{1079}\\\hline
 \multicolumn{1}{|c|}{Testing}  & \numprint{606} & \numprint{1964} & \numprint{532} & \numprint{1513} & \numprint{577}\\\hline
\end{tabular}
\end{scriptsize}
\caption{Number of images for each category in the training and testing sets. There are \numprint{11380} training and \numprint{5192} testing images}
\label{tab:n_images}
\end{table}

Following~\cite{Wu2009} we define the transition matrix as $\condprob{\state_{t}}{\state_{t-1}} = p_e$ if $\state_t=\state_{t-1}$;
the rest of the probability mass is shared uniformly among all other transitions.
We use $p_e = 0.99$.

In order to test the influence of the $n$-gram order we have varied $n$ between 1 and 6.
Similarly we have tested the Lidstone-Laplace (LL) smoothing with parameter $\delta = 1$ and the Witten-Bell (WB) smoothing.
The training set was too small to use the Knesser-Nay smoothing. 
In our experiments we use interpolated models~\cite{Manning1999}.
We have tested several values of the sub-sampling rate: $\sampleRate=1$ (which has no effect), $\sampleRate=3$ and $\sampleRate=5$.
We use $\compressRate=3$ for the ``compress'' strategy.
The ``unique'' strategy don't need any parameter.

Setting $n=1$ with Lidstone-Laplace smoothing gives the same temporal integration method than in~\cite{Wu2009}
(note that we don't use the same signature).
Setting $n=2$ with Lidstone-Laplace smoothing and without interpolation gives a system similar to~\cite{Dubois2011}.

\subsection{Results}

\begin{sidewaysfigure*}
 \setcounter{subfigure}{0}
 \centering
 \subfigure[$\somSize=10$, Lidstone-Laplace smoothing]{\includegraphics[width=0.45\textwidth]{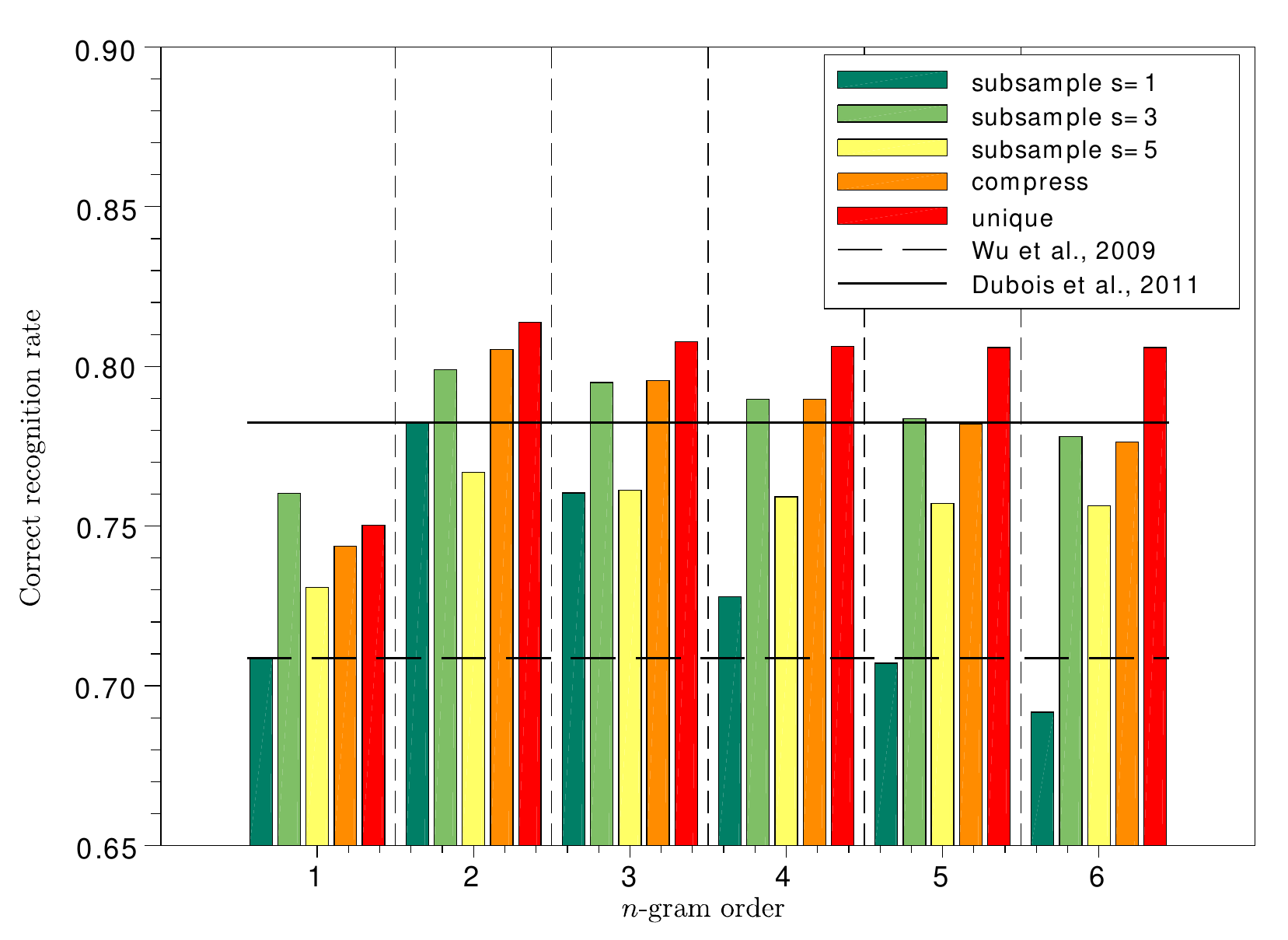}\label{fig:results_gist_gray_10_ll}}
 \subfigure[$\somSize=10$, Witten-Bell smoothing]{\includegraphics[width=0.45\textwidth]{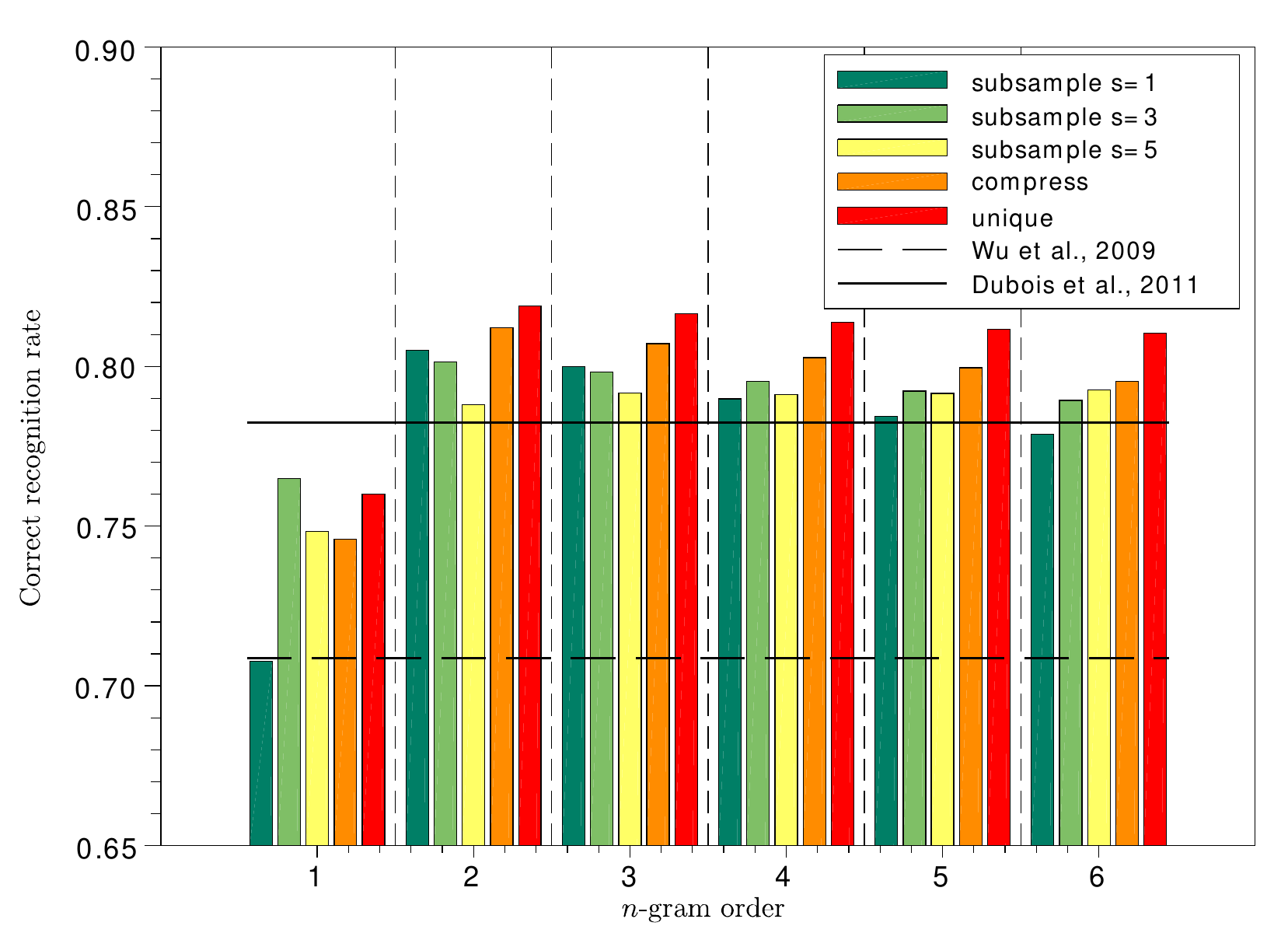}\label{fig:results_gist_gray_10_wb}}

 \subfigure[$\somSize=20$, Lidstone-Laplace smoothing]{\includegraphics[width=0.45\textwidth]{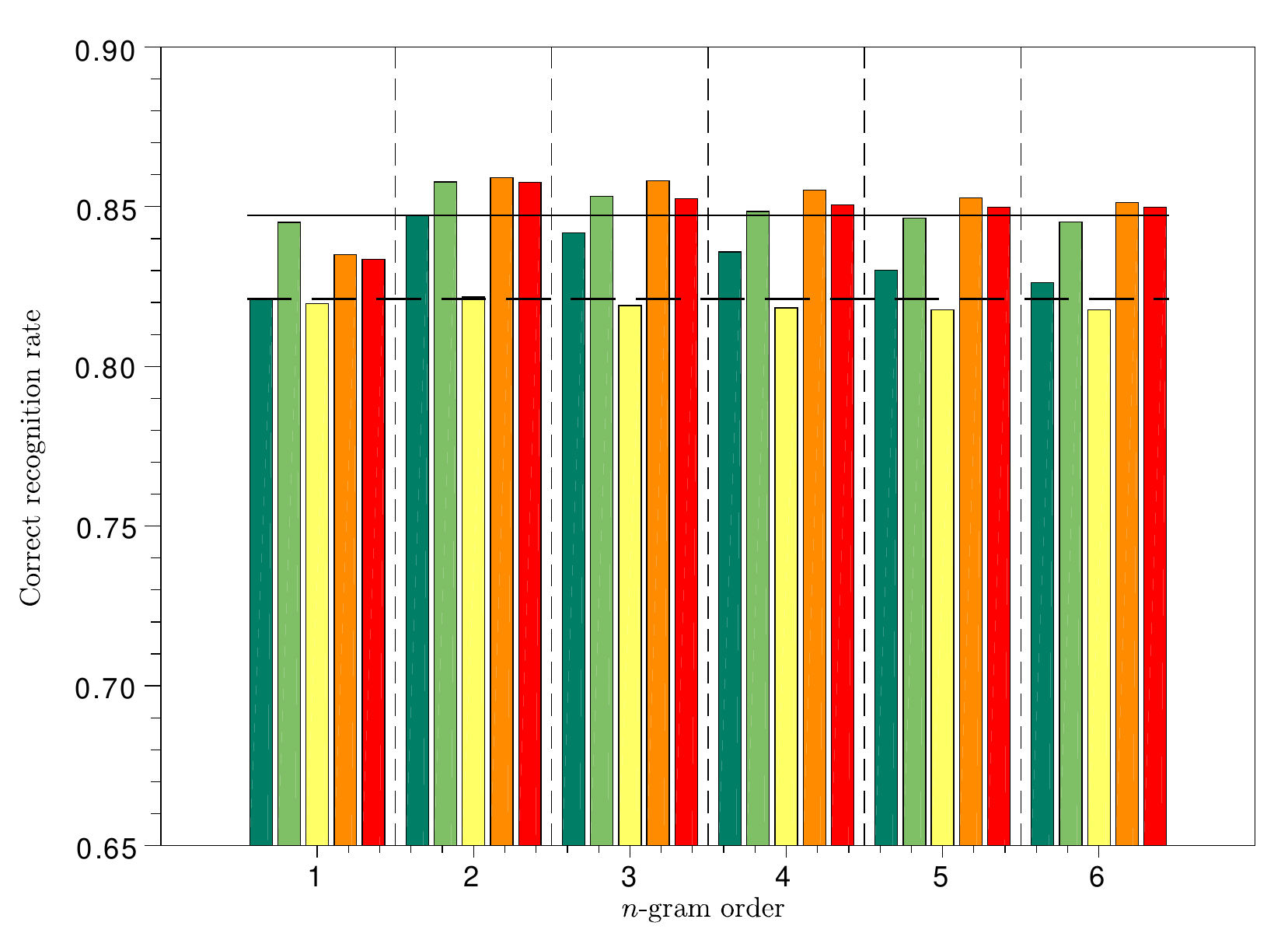}\label{fig:results_gist_gray_20_ll}}
 \subfigure[$\somSize=20$, Witten-Bell smoothing]{\includegraphics[width=0.45\textwidth]{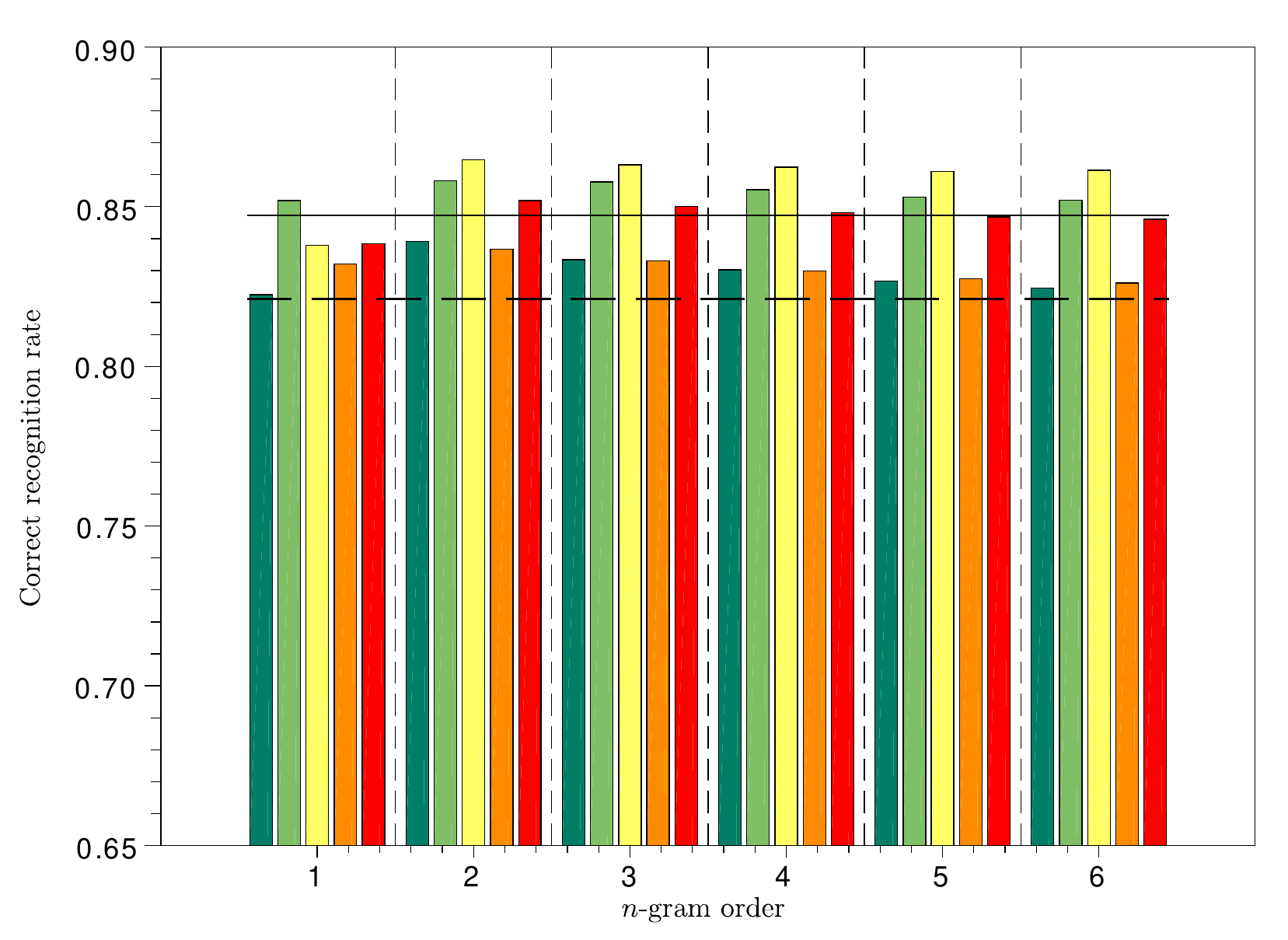}\label{fig:results_gist_gray_20_wb}} \\
 \caption{Results on the instance recognition task.
 The vertical axis is the correct recognition rate (in \%). The horizontal axis is the value of $n$.
 Upper-row: results for $\somSize = 10$. Lower row: results for $\somSize = 20$. Left column: results for Lidstone-Laplace smoothing. Right column: results for Witten-Bell smoothing.
 }
 \label{fig:results_gist_gray}
\end{sidewaysfigure*}

Results are presented on figure~\ref{fig:results_gist_gray}.
It must be noted that on this instance recognition task, a larger~\ac{SOM} gives better results.
This is expected from the literature (see~\cite{Guillaume2011}).
The second observation that could be made is that the word selection methods generally increase the results by several percent.
This can be seen by the difference between the bar for $\sampleRate=1$ and other bars of the same group.
The ``subsample'' strategy with $\sampleRate=3$ is rather efficient, sometimes increasing performance by 6\%.
Setting $\sampleRate=5$ generally gives less important increase.
Performance decreases with $\somSize=20$ and LL smoothing.
However, this strategy leads to the best results on the task for $\somSize=20$ and WB smoothing.
The ``compress'' strategy is usually efficient except for $\sampleRate=20$ and WB smoothing.
The ``unique'' strategy is always among the best choices and it's results are less sensitive to the $n$-gram order.
Generally speaking, the effect of those strategies increase with $n$.
Results with WB smoothing are generally a little bit better than with LL in particular for large $n$.

It is clear from the figure that using $n=2$, \ie{} to take into account the dependence on the last image, is a clear improvement over $n=1$,
\ie{} the classical \ac{HMM}.
However using $n$-grams with $n>2$ has little impact on performance.
It should be noted that when $\sampleRate=1$, the performance drops when $n>2$.
With word selection, the performance can be high with large $n$.
This seems to confirm the intuition behind the word selection techniques.

\section{\uppercase{Conclusion}} \label{sec:conclusion}

\noindent We have presented a new model of temporal integration using \ac{HOHMM} for semantic place recognition which
models the dependence between observations.
We have shown that taking this dependence into account can lead to interesting gains in performance.
However, contrary to what we expected, using larger $n$ don't improve performance.
The smoothing technique seems to have minor effect.
This may be caused by the fact that we use relatively small training sets compared to the field of~\ac{NLP}
where those techniques have been developed.
Those results must take into account the fact that recognition rates are already quite high on the task studied here.

We have shown that simple methods to select important words could improve the results.
Our results suggest that large $n$ could be interesting if combined with good word selection techniques.

Future works will focus on the vector quantization process to learn better words.
More sophisticated word selection techniques may also be useful.
Finally we could also look for more discriminative descriptors.

\section*{\uppercase{Acknowledgements}}
\noindent We thanks Thiago Fraga and Alexandre Allauzen for fruitful discussion and help with the SRILM toolkit.

\bibliographystyle{apalike}
{\small
\bibliography{\bibpath/bib_visapp2013}}

\begin{thebibliography}{}

\bibitem[Berchtold, 2002]{Berchtold2002}
Berchtold, A. (2002).
\newblock High-order extensions of the double chain markov model.
\newblock Technical Report 356, University of Washington.

\bibitem[Chen and Goodman, 1996]{h_Chen1996}
Chen, S.~F. and Goodman, J. (1996).
\newblock An empirical study of smoothing techniques for language modeling.
\newblock In {\em Proceedings of the 34th annual meeting on Association for
  Computational Linguistics}.

\bibitem[Dubois et~al., 2011]{Dubois2011}
Dubois, M., Guillaume, H., Tarroux, P., and Frenoux, E. (2011).
\newblock Visual place recognition using bayesian filtering with markov chains.
\newblock In {\em Proceedings of the European Symposium on Artificial Neural
  Networks (ESANN 2011)}.

\bibitem[Filliat, 2008]{Filliat2008}
Filliat, D. (2008).
\newblock Interactive learning of visual topological navigation.
\newblock In {\em Proceedings of the 2008 IEEE International Conference on
  Intelligent Robots and Systems (IROS 2008)}.

\bibitem[Guillaume et~al., 2011]{Guillaume2011}
Guillaume, H., Dubois, M., Tarroux, P., and Frenoux, E. (2011).
\newblock {Temporal Bag-of-Words: A Generative Model for Visual Place
  Recognition using Temporal Integration}.
\newblock In {\em Proceedings of the International Conference on Computer
  Vision Theory and Applications (VISAPP 2011)}.

\bibitem[Kohonen, 1990]{h_Kohonen1990}
Kohonen, T. (1990).
\newblock The self-organizing map.
\newblock In {\em Proceedings of the IEEE}, volume~78, pages 1464--1480.

\bibitem[Lee and Lee, 2006]{Lee2006}
Lee, L.-M. and Lee, J.-C. (2006).
\newblock A study on high-order hidden markov models and applications to speech
  recognition.
\newblock In Ali, M. and Dapoigny, R., editors, {\em Advances in Applied
  Artificial Intelligence}, volume 4031 of {\em Lecture Notes in Computer
  Science}.

\bibitem[Manning and Sch{\"u}tze, 1999]{Manning1999}
Manning, C. and Sch{\"u}tze, H. (1999).
\newblock {\em Foundations of statistical natural language processing}.
\newblock MIT Press.

\bibitem[N\"{u}chter and Hertzberg, 2008]{Nuchter2008}
N\"{u}chter, A. and Hertzberg, J. (2008).
\newblock Towards semantic maps for mobile robots.
\newblock {\em Robotics and Autonomous Systems}, 56(11):915--926.

\bibitem[Pronobis and Caputo, 2007]{Pronobis2007}
Pronobis, A. and Caputo, B. (2007).
\newblock Confidence-based cue integration for visual place recognition.
\newblock In {\em Proccedings of the 2007 IEEE/RSJ International Conference on
  Intelligent Robots and Systems}.

\bibitem[Pronobis et~al., 2010]{Pronobis2010}
Pronobis, A., Mozos, O.~M., Caputo, B., and Jenseflt, P. (2010).
\newblock Multi-modal semantic place classification.
\newblock {\em The International Journal of Robotics Research},
  29(2-3):298--320.

\bibitem[Ranganathan, 2010]{Ranganathan2010}
Ranganathan, A. (2010).
\newblock {PLISS}: Detecting and labeling places using online change-point
  detection.
\newblock In {\em Proceedings of the 2010 Robotics: Science and Systems
  Conference (RSS 2010)}.

\bibitem[Torralba et~al., 2003]{h_Torralba2003}
Torralba, A., Murphy, K.~P., Freeman, W.~T., and Rubin., M.~A. (2003).
\newblock Context-based vision system for place and object recognition.
\newblock In {\em Proceedings of the Nineth IEEE International Conference on
  Computer Vision (ICCV 2003)}, volume~1, pages 273--280.

\bibitem[Ullah et~al., 2008]{Ullah2008}
Ullah, M.~M., Pronobis, A., Caputo, B., Luo, J., Jensfelt, P., and Christensen,
  H.~I. (2008).
\newblock Towards robust place recognition for robot localization.
\newblock In {\em Proceedings of the IEEE International Conference on Robotics
  and Automation (ICRA 2008)}, Pasadena, USA.

\bibitem[Vasudevan et~al., 2007]{Vasudevan2007c}
Vasudevan, S., Gachter, S., Nguyen, V., and Siegwart, R. (2007).
\newblock Cognitive maps for mobile robots--an object based approach.
\newblock {\em Robotics and Autonomous Systems}, 55(5):359--371.

\bibitem[Wu et~al., 2009]{Wu2009}
Wu, J., Christensen, H., and Rehg, J. (2009).
\newblock Visual place categorization: Problem, dataset, and algorithm.
\newblock In {\em IEEE/RSJ International Conference on Intelligent Robots and
  Systems, 2009 (IROS 2009)}.

\end{thebibliography}


\vfill
\end{document}